\def\year{2023}\relax
\documentclass[letterpaper]{article} % DO NOT CHANGE THIS

\usepackage{aaai21FORPREPRINTONLY} 

\usepackage{times}  % DO NOT CHANGE THIS
\usepackage{helvet} % DO NOT CHANGE THIS
\usepackage{courier}  % DO NOT CHANGE THIS
\usepackage[hyphens]{url}  % DO NOT CHANGE THIS
\usepackage{graphicx} % DO NOT CHANGE THIS
\usepackage{natbib}  % DO NOT CHANGE THIS AND DO NOT ADD ANY OPTIONS TO IT
\usepackage{caption} % DO NOT CHANGE THIS AND DO NOT ADD ANY OPTIONS TO IT
\title{Beyond Discrete Genres: \\Mapping News Items onto a Multidimensional Framework of Genre Cues}
\author {
    Zilin Lin\textsuperscript{\rm 1},
    Kasper Welbers\textsuperscript{\rm 2},
    Susan Vermeer\textsuperscript{\rm 1},
    Damian Trilling\textsuperscript{\rm 1} \\
}
\affiliations {
    \textsuperscript{\rm 1} University of Amsterdam \\
    \textsuperscript{\rm 2} Vrije Universiteit Amsterdam
}

\begin{document}

\maketitle

\begin{abstract}
In the contemporary media landscape, with the vast and diverse supply of news, it is increasingly challenging to study such an enormous amount of items without a standardized framework. Although attempts have been made to organize and compare news items on the basis of news values, news genres receive little attention, especially the genres in a news consumer's perception. Yet, perceived news genres serve as an essential component in exploring how news has developed, as well as a precondition for understanding media effects. We approach this concept by conceptualizing and operationalizing a non-discrete framework for mapping news items in terms of genre cues. As a starting point, we propose a preliminary set of dimensions consisting of ``factuality'' and ``formality''. To automatically analyze a large amount of news items, we deliver two computational models for predicting news sentences in terms of the said two dimensions. Such predictions could then be used for locating news items within our framework. This proposed approach that positions news items upon a multidimensional grid helps in deepening our insight into the evolving nature of news genres.
\end{abstract}

\noindent How to compare a newspaper article on elections with a celebrity interview on YouTube, if they are both broadly defined as news? In today's media landscape, the public is updated on current affairs through various routes \cite{kaul2012changing}. With an ever-expanding notion of news, the scholarly challenge is to systematically study this wide range of content. Moving beyond classic cross-outlet observations \cite{riffe1997content} and drawing comparisons on other levels (such as a cross-context one between newspaper pieces and search engine results, or a cross-medium one between public broadcasts and YouTube channels) could not only help with investigating how news has diverged, but also serve as a first step to understanding media effects. It is, however, becoming increasingly difficult to build a unified framework for such comparisons, due to the evolving nature of news \cite{vraga2016blurred,yadamsuren2011online}. This study therefore asks: How can a vast and diverse amount of news items be mapped onto one standardized framework, representing the variety of what constitutes news?

Attempts have been made to organize news items along abstract dimensions. One approach zooms in on news events \cite{trilling2020}. For instance, research in this vein focuses on how certain events are selected by traditional gatekeepers \cite{gans1979deciding,harcup2001news}, or whether such selection criteria (i.e., news values) are comparable to the ones applied by other actors (e.g., news consumers) in the context of online dissemination \cite{harcup2017news,trilling2017newsworthiness}. Alternatively, the genre approach distinguishes news from a narrative perspective \cite{buozis2018reading}. This type of analysis, though being extensively applied within the disciplines of rhetoric, literary, and films \cite{mittell2001cultural}, has rarely been performed in the news domain \cite{buozis2018reading}. Essentially, it aggregates news items by their stylistic representation \cite{dent2008journalists}, resulting in different categories, such as feature, commentary, and editorial. 

It should, nonetheless, be noted that, addressing the evolving journalistic practices today \cite{bakhtin1986speech}, genres are not rule-governed nor static, but ``fluid and dynamic'' \cite[p. 10]{freedman1994genres}. This nature of news genres, thus, does not harmonize with a constant taxonomy. As a preliminary solution, the present study moves beyond discrete genre demarcations and puts forward a multidimensional framework where news items could be positioned in a continuum, representing whether and how boundaries of news genres are shifting and blending. This provides a novel understanding of news genres in the contemporary media landscape where they are not as clearly separated as they were by legacy media but gradually blurring.

Moreover, instead of being intrinsic to the text \cite{mittell2001cultural}, genres are defined by how the text and its context interact \cite{schryer2005genre}. In news communication, a typical context involves news consumers' interpretation \cite{mittell2001cultural}. Accounting for this, we position news items on the basis of genre cues, the observable features that news consumers use to recognize genres \cite{kessler1997automatic}. Specifically, they are manifested in the linguistic features of a news item. We choose two dimensions as a starting point, namely ``factuality'' and ``formality''. Being different from previous research, which has been newsroom-centered, our proposed framework helps understanding whether and how the audience distinguishes news genres \cite{visch2008narrative}. 

To summarize, in this paper, we conceptualize and operationalize a two-dimensional framework for mapping news items in terms of genre cues. For the automatic positioning, we fine-tuned two sentence-level BERT models with a large and varied annotated corpus that comprises news sentences from diverse Dutch media outlets, including blog posts, online newspapers, podcasts, public broadcasts for different age groups, satirical programmes, and YouTube channels. We then showcased how our framework could be utilized to capture the developments of news genres using large-scale data. The showcases, in the meantime, validated our approach. As discussed, this paper contributes both theoretically and methodologically. On the societal level, it could also be beneficial when evaluating news products, further helping practitioners improve their journalistic decisions.

\section{Theoretical Framework}

\subsection{Journalistic Genres and the Traditional Genre Approach}

Human activity involves the use of language --- participants of an activity realize the language in the form of utterances, reflecting in thematic content, linguistic style, and compositional structure \cite{bakhtin1986speech}. Genres are the relatively stable types of these utterances developed in different communicative spheres \cite{bakhtin1986speech}. From a pragmatic perspective, genre is seen as a rhetorical action that recurs in a particular social environment \cite{miller1984genre}, or as \citet{swales1990genre} defines it, a discourse community (i.e., a group with goals and a shared set of discourses for communication). In this paper, when studying journalistic texts, taking the genre approach means to find ``the linguistic features aggregated in recognizable patterns'' \cite[p. 1]{giltrow2009genres}. Analyzing these aggregations provides insights into the complex nature of news communication.

In journalistic practice, this notion of genre has been commonly applied, with exemplars like breaking news, investigative reporting, and commentary, as listed by the Pulitzer Prizes \cite{pulitzer2022}. In academia, in contrast, the concept of genre has not drawn much attention of communication scientists \cite{broersma2018exploring}. Most scholars working on such topics have attempted to distinguish different journalistic genres, defining them as ``means of expressions sought by journalists'' \cite[p. 11]{gargurevich1982genre}, or ``the various stylistic methods applied by professionals'' \cite[p. 51]{albertos2004genre}. The distinguished genres have been further classified by functions (e.g., information, guidance, interpretation, entertainment, etc.) and examined in empirical studies \cite[e.g.][]{colussi2020examining, melo2016journalistic, seixasndfurther}. Instead of differentiating pre-established genres in newsrooms, another group of researchers have analyzed published news items via the genre approach, famously categorizing them as ``hard news'' or ``soft news'' \cite{lehman2010hard}. These two concepts have elicited a fruitful discussion in terms of news classifications based on news topics \cite{lumby1994feminism}, news styles \cite{patterson2000doing}, and news mediums \cite{baum2003soft}\footnote{It should be noted that the distinction between ``hard news'' and ``soft news'' involves both ``topical genres'' concerning news topics or ``beats'' and ``modal genres'' that refer to formal conventions \cite{hartsock2002genre}. As mentioned before, this study mainly focuses on ``modal genres''.}. 

As discussed above, the existing literature in this field shares a top-down approach that begins with distinguishing a fixed set of genres \cite[p. 4]{giltrow2009genres}. In this case, according to \citet*[p. 5]{giltrow2009genres}, the onus for genre identification is on the theorist to position a closed set and on the analyst to classify items. Although this approach has generated some knowledge in terms of journalistic genres, it suffers from two main issues, on which we will elaborate in the following two sections.  

\subsection{Moving beyond Discrete Genres}

The first problem of the traditional top-down approach is that it fails to capture how news genres evolve in today's media landscape, even though it allows static classifications of a large amount of news items. According to genre theories, the use of language is as complex as the activity it involved \cite{bakhtin1986speech}. In the modern news sphere, the complexity is evident: the public is now updated on current affairs through an overwhelming amount of information from countless channels, and the notion of news has been drastically changed without consensus being reached \cite{kaul2012changing}. Therefore, as journalistic practice \cite{yates1992genres}, news genres are ``in a constant process of negotiation and change'' \cite[p. 137]{buckingham1993children}. It is then unrealistic to expect a fixed set of pre-defined genres illuminating such dynamic constructions and their overtime development \cite{broersma2018exploring}.

The bottom-up approach, on the contrary, opens up the possibility for new genres to emerge. In the era of legacy media, established news genres serve as a ``formal provision'' that rigidly regulates journalistic activities \cite[p. 1077]{serdali2016genres}. However, the guiding role of genres can no longer be assumed in the contemporary journalistic environment \cite{artemeva2009stories}, especially considering the rise of non-institutionalized practitioners. In fact, researchers have suggested that genre definition is only possible post factum \cite{serdali2016genres}. The bottom-up approach that identifies patterns among news texts is thus deemed appropriate to study news genres in flux, resulting in an open list of specific genres \cite{giltrow2009genres}. That being said, one concern regarding this approach is that it could lead to a multitude of genres on an open yet ungeneralizable list \cite{giltrow2009genres}, on which, for instance, a single news item might constitute one genre. When studying abundant news content, this list of lower-level genres might be too trivial to provide generalizable insight into the news realm as a whole. On this account, news genres should better not be seen as separate units --- isolating one genre from another and looking into the typology per se are by no means the ideal ways of classification. 

According to genre theories, genres are non-discrete systems without rigid rules, confronting us with the challenge of delimitation \cite{gledhill1985genre}. Mixed genres are common among mass media products \cite[p. 89]{fairclough1995media}. In the news domain, specifically, the multi-purpose journalistic practice results in the genre heterogeneity of news \cite{van1993genre}. The fuzzy ``news-ish'' products are instantiations of a hybrid journalistic purpose, such as to inform and to discuss, or update and to entertain \cite{edgerly2017making}. Infotainment, as a genre with mixed functions \cite{otto2017softening}, is a typical example. Furthermore, digital media have led to new hybrid news genres \cite{colussi2020examining}, which consequently turn into the rule rather than exceptions \cite{mast2017hybridity}. With the boundaries being blurry and permeable, news genres are becoming undefinable though most of the time still recognizable \cite{chandler1997introduction}.

News genres are constantly evolving over time \cite{chandler1997introduction}: They mix and muddle when  ``old'' genres fade away and when ``new'' genres emerge \cite{seixasndfurther}. Yet, changes in genres are rather gradual and ``familiarity remains dominant'' \cite[p. 276]{ekstrom2002epistemologies}. Moreover, when it comes to genre terms, news genres appear to be relatively stable in formations (e.g., satire, op-ed, etc.), whose dynamic definitions are agreed by the majority of the public \cite{mittell2001cultural}. Thus, to study this ``stability in flux'', and to capture both consistencies and variations, one solution is to move on from discrete genre elements and realize news genres as stable clusters from a broader scope \cite[p. 11]{mittell2001cultural}.

Taken together, acknowledging the fluidity and impurity of news genres, we propose to study them as evolving clusters in a continuum \cite{mittell2001cultural}. Moving beyond distinct genre demarcations, we propose to map news items onto a multidimensional framework, which allows the emergence and aggregations of genre clusters, and more importantly, reveals how news genres have been shifting and blending across time and/or space (e.g., domains, outlets, etc.). 

\subsection{Genre Cues and Linguistic Features as Proxy}

The other issue with the traditional genre approach is its over-reliance on the ``supply-side'' of the journalistic language. Genres are situational and they do not merely amount to the text itself \cite{biber1989drift}. Based on \citet*{miller1984genre}'s definition, the loci for studying genres should be both textual attributes and cultural practices that constitute the texts, such as production and interpretation \cite{fiske2002television,mittell2001cultural}. Specifically, to situate news texts within journalistic practices, genres could be defined as the tacit knowledge mediating between news producers and news users \cite{chandler1997introduction,feuer1992genre,mcquail1987mass,tolson1996mediations}. Yet, the discussion till now has been overtly newsroom-oriented \cite{costera2020understanding}. Given the crucial and gradually active role the audience plays in nowadays' journalistic practice \cite{cohen2019work,ferrer2018audience}, the present study acknowledges that the onus for genre recognition is as well on the ``ordinary language users'' \cite[p. 5]{giltrow2009genres}, thereby suggesting a new genre approach aligning with the audience's perception.

News genres are equally crucial to the audience: they not only refer to certain journalistic styles and convey the identity of news producers \cite{broersma2018exploring}, but also serve as a ``functional entity'' that provides news users a sense of orientation when consuming the news subject \cite[p. 1083]{serdali2016genres}. In other words, genres package contextual information that in turn instructs the audience to deploy their corresponding processing strategies and knowledge types \cite{giltrow2009genres}. The audience delimits news genres through certain mechanisms. In a relatively traditional news context where the public update themselves through news institutions, audiences recognize the genres from ``pre-signals'' \cite[p. 4]{giltrow2009genres}, such as the classification tag of an article on the news website, the position of different pieces within the newspapers, the number of guest speakers present in the news television programme, etc. Specifically for news consumers, an ``advertisement'' tag could properly alert them to the taken information on the news website, a square of text on the newspaper's opinion page may automatically be perceived as subjective, and when there is no guest speaker but a suit-up anchor, the programme being news broadcasting would be a plausible guess. With these ``contextual indicators'' \cite[p. 1156]{broersma2018exploring}, textual cues might not necessarily be engaged.

However, in the face of the blooming social media, users are exposed to numerous options of news information, where the aforementioned ``pre-signals'' are becoming less and less explicit, even being completely removed. Although it is difficult for the audience to articulate the ground for delimitation \cite{chandler1997introduction}, there are indeed other cues deployed for inference, being related to different components of a news item, which include styles of the text \cite{buozis2018reading}. In the present study, we conceptualize genre cues as such styles \cite{langholz1987mass}. As our genre approach intends to study news items as if they would be perceived by the audience on the basis of these genre cues, we manifest this concept in the observable textual properties of a news item that the audience uses to distinguish genres \cite{kessler1997automatic}, typically the linguistic features.

The reasons for using linguistic features as proxy for genre cues are as follows. Grounded in genre theories, it is the surface enunciation, instead of the deep meaning beneath the textual surface, considered the best property when analyzing genre discourses \cite{mittell2001cultural}. \citet*{swales1990genre} argues that the rationale for genres, constituted by communicative purposes, is in close relation to the usage of linguistic styles. Similarly, \citet*{bakhtin1986speech} views linguistic cues as the emphasis of genre recognition. We, therefore, choose linguistic cues as the articulation of news genres, which has not been a common practice in the field of news communication research where news content has long been preferred over styles \cite{harcup2001news}. Considering the linguistic ``fusion'' in individual news genre \cite[p. 11]{giltrow2009genres}, the interplay of multiple stylistic dimensions could be safely assumed. Taking the genre approach, since the key is to focus on the breadth of surface articulation rather than the depth of texts \cite[p. 45]{edgerton2005thinking}, this study chooses the following two classic journalistic dimensions as a starting point, and treat them as a continuum in our multidimensional framework.

\subsection{A Starting Point: The Good Old Dimensions}
With the golden rules of being factual and formal, news genres have been established accordingly in history. For example, an opinion piece denotes news with a fair amount of subjective content while public broadcasts would be expected to be more formal than a satirical programme. Today, with the notion of news becoming ever-inclusive and journalistic practices being accessible to not merely the professionals, news does not necessarily adhere to the traditional format, appearing more than ``boring and grey'' \cite[p. 2333]{costera2020understanding}. It is, then, of importance to explore to what extent news today has deviated from these two classic journalistic norms.

\subsubsection*{Factuality.}
Presenting only facts without personal involvement has been key in journalistic practice \cite{schudson2001objectivity}. Yet, in this digital era, catering to particularly the functional social media logic of popularity and connectivity \cite{van2013understanding}, the objective\footnote{While we acknowledge that the notion of objectivity does not equal to being factual, here the term is mainly used to denote the density of factual information in a news item.} value of a news item has been greatly negotiated. Opinion-orientated items could not only boost dissemination \cite[e.g.,][]{peer2011youtube} but also prompt engagement, which help in building a more dialogic form of news communication \cite{singer2005political}. Naturally, personal frames and subjective references are now frequently featured in the news domain \cite{blake2019news}, and objective reporting has been especially challenged on social media: \citet*{lasorsa2012normalizing} noticed that this professional norm of avoiding opining was undermined on Twitter --- journalists' tweets convey a substantial amount of opinions; and on Facebook, even newspapers use subjective language \cite{welbers2019presenting}. Distinctively, citizen journalists (e.g., YouTubers) viewed this subjectivity as their edge on opposing new organizations and advocated personal perspectives when distributing news for the attention-gaining purpose \cite{lewis2018alternative,marwick2015you}. 

\subsubsection{Formality.}
Compared with the first dimension, far too little attention has been paid to (in)formality in the news domain. The main reason is that in theory, and more importantly, in everyone's impression, news is always associated with a formal style --- even when constructing a formality corpus, news text was directly used by computational scientists as the formal data \cite{sheikha2010automatic}. However, the statement that news  ``ought not to be fun'' \cite[p. 107]{costera2007paradox} has been contested from both journalists and the audience's points of view: during the shift towards the ``audience turn'', the effort to present news in an appealing format has been witnessed in news production \cite{costera2007paradox}; while the audience today, due to the revolutionary technological changes, is updated on public issues not only in a traditional way from serious newspapers but from different types of channels with diverse formats, such as satire or a YouTube video that goes viral. Given that the essential function of news does not vary with its format \cite{costera2007paradox}, even though news was consistently reported as one the most formal registers in the previous literature \cite{lahiri2015squinky}, the present study suggests that it is of relevance to be more inclusive, exploring the subtle layers of formality among various forms of items that engage the audience with news events.

Despite the recent rise of alternative news along these two dimensions (opinion-oriented and/or informal news), for news consumers, being factual and being formal remain the central indicators of ``real news'' \cite{yadamsuren2011online}. What they do not agree on, however, is the understanding in these two dimensions \cite{pavlick2016empirical}. For instance, the audience does not always perceive facts and opinions in the same way as their professional counterpart, while (in)formality is as well an inherently subjective concept, reaching poor consistent upon the notion of being (in)formal \cite{lahiri2011inter,lahiri2011informality,lahiri2015squinky}. 
\subsection*{The Framework}
To recapitulate, considering the fluid and contextual nature of news genres, instead of studying merely the text itself or clear-cut genre demarcations, we propose to position news items on a multidimensional grid. For the purpose of this paper, we limit ourselves to the two dimensions discussed in detail above: factuality and formality. Through exploring how news genres diverge and converge along these two dimensions of genre cues, comparisons of news items on both individual and aggregate levels are allowed. Specifically, positioning the aggregated categories on such a grid could add extra insight into how news genres appear differently among topics, outlets, or even time periods. 

We develop a method to realize this way of systematization automatically. To showcase the performance, and to validate our proposed framework, we apply this tool to capture how traditionally discrete genres are mapped onto the two-dimensional grid with Dutch news-related items from different sources.

\section{Method}
This section introduces the steps towards building a tool for quantifing news items on the scales of factuality and formality. We first collect a diverse corpus, which after crowd-sourced annotation is used as the input for model training. 

\subsection{Data Acquisition and Preparation}

\begin{table}[t]
\centering
%\resizebox{.95\columnwidth}{!}{
\begin{tabular}{l|l|c}
    & Outlet & \textit{N}\\
    \hline
    Blog posts & \textit{Geenstijl} & 404\\
    & \textit{Stukroodvlees} & 524\\
    & \textit{Tweakers} & 636\\
    \hline
    Opinion articles & \textit{NRC Handelsblad} & 354\\
    & \textit{De Volkskrant} & 319\\
    & \textit{Trouw} & 169\\
    & \textit{De Telegraaf} & 152\\
    \hline
    Podcasts & \textit{Boekestijn en De Wijk} & 614\\
    & \textit{De Correspondent} & 684\\
    & \textit{Maarten van Rossem} & 607\\
    \hline
    TV satire & \textit{Zondag met Lubach} & 1,862\\
    \hline
    TV broadcasts & \textit{Jeugdjournaal} & 8,001\\
    & \textit{NOS Journaal} & 3,355\\
    \hline
    YouTube & \textit{NOS op3 uitgediept} & 1,022\\
\end{tabular}
\caption{An overview of the corpus (18,703 sentences in total, 2009-2022).}
\label{table1}
\end{table}

In an attempt to build a suitable corpus for studying both dimensions (i.e., factual vs. opinion-oriented and formal vs. informal), we collected 214 news items from a diverse set of Dutch news outlets, including written-text data from newspapers online and blog posts, and spoken-text data (i.e., subtitles and transcripts) from television programs, YouTube videos, and podcasts. All items were then tokenized into sentences and we only kept sentences with a ``regular'' length (i.e., between 5 and 50 words). Also, sentences containing words that indicate the source (e.g., ``Jeugdjournaal'', ``Lubach'', etc.) were removed to avoid leaking information during model training. These pre-processing steps leave us with 18,703 sentences in the final corpus (see Table \ref{table1}).

\subsection{Annotation and Post-processing}
In order to estimate the news consumers' perception of each sentence, we recruited 17 Dutch native speakers for the annotation task\footnote{Crowd-sourcing on MTurk was the initial plan. However, the pilot annotation task with a customized qualification test required showed that there are not enough active Dutch Turkers to finish this task in a short term. We therefore adopted this alternative approach and retained the qualification test.}. The annotation procedure lasted for seven weeks in total from 26/08/2021 till 13/09/2021 (batch 1) and from 08/04/2022 till 29/04/2022 (batch 2). After passing the customized qualification tests, annotators were recruited and provided with the coder instructions and an unlabelled data set for the assignment. In the coder instructions, we specifically asked the annotators to judge the sentence as a sentence \textit{in the news domain}. Both dimensions are discussed in the next paragraphs.

\subsubsection{Factual vs. Opinion-oriented.}  \citet*{alhindi2020fact} have shown that argumentation features are promising when distinguishing news stories from opinion articles: To prepare the input for document classification, a model was trained to predict the argumentation features for each sentence using a corpus with fine-grained argumentative discourse unit annotations from \citet*{al2016news}, during which they grouped the original argumentative features into three coarser types, namely claim (assumption), premise (common ground, testimony, statistics, anecdote), and other (other) \cite{alhindi2020fact}. With the purpose of estimating news consumers' perception instead of obtaining professionally-defined features, we transformed this approach into a more laymen-friendly version by asking the annotators to choose among ``fact'', ``(also) opinion'', and ``neither of them''. The annotators were instructed by a simple rule that a ``fact'' sentence ``states factual information that you believe is either accurate or inaccurate'' while an ``opinion'' sentence ``contains some forms of subjective content that you either agree or disagree with'' \cite{mitchell2018distinguishing}.

\subsubsection{Formal vs. Informal.} The common practice in formality annotation is to ask for intuitions \cite{lahiri2011inter,lahiri2015squinky}, which is perfectly aligned with our goal --- understanding how news consumers perceive the sentences. After receiving the poorly-agreed annotations, \citet*{lahiri2011informality} suggested that adopting a Likert scale and merging the categories after instead of using a binary answering format could be one of the solutions for agreement improvement. We followed this advice and adjusted \citet*{park2014identifying}'s measurement by providing the annotators with a 5-point answering scale, ranging from ``very informal'' to ``very formal''.

Although using the same corpus, the two annotation tasks were separately assigned to avoid potential bias (e.g., one might label a sentence as ``fact'' when it appears to be ``formal''). For both tasks, each sentence was labeled by three annotators, who achieved an inter-coder agreement (i.e., Krippendorff's $\alpha$) of 0.52 for the fact-opinion task and 0.47 for the formal-informal task. It should be noted that although this seems insufficient, these two reliability scores: a). were comparable to the levels of agreement in the previous studies \cite{al2016news,lahiri2011informality,park2014identifying}; b). were expected as we did not ask for the ``correct'' labels but intuitions on subjective tasks from random news consumers without training beforehand; c). should be seen as only a reference given the fact that there were 17 annotators dividing the tasks.

To guarantee accuracy, we applied the rule of majority votes for deciding the final labels for each sentence, during which the sentences with tied votes were discarded. We also removed the ``neutral'' category in the formality task due to its inadequate amount. In the end, there are two corpora with 18,310 (69.28\% fact, 23.40\% opinion, and 7.31\% neither) and 17,387 (40.72\% informal and 59.28\% formal) sentences prepared for a three-way classification and a two-way classification respectively.

\subsection{Model Training}

Following \citet*{alhindi2020fact}, we chose the BERT model for sentence classification in both tasks \cite{devlin2018bert}. To be more context-specific, a monolingual Dutch model (i.e., BERTje) that based on a diverse dataset containing three news corpora was chosen over the original one that only based on Wikipedia text \cite{de2019bertje}. Before the model training, we split the data into a training set (80\%) and a test set (20\%), and the training set was further split into a training set (80\%) and a validation set (20\%). For both tasks, we experimented the models with different combinations of hyper-parameters. In addition, as reference, we trained three baseline models (i.e., Na\"{\i}ve Bayes, Logistic Regression, and Support Vector Classification) with two different vectorizers (i.e., CountVectorizer and TfidfVectorizer) and presented the results for comparisons after hyperparameter optimization using GridSearch.

\section{Results}
We first discuss the performance of different models. After having established that the performance is sufficient, we present two case studies to illustrate how communication scientists can apply this method.

\subsection{Model Performance}
In both tasks (i.e., factual vs. opinion-oriented vs. neither and formal vs. informal), as shown by Table \ref{table2}, the BERT models perform evidently better than the other three baseline models, regardless of different combinations of hyper-parameters, with which the models yield negligible differences in their performance. 

\begin{table}[t]
\centering
%\resizebox{.95\columnwidth}{!}{
\begin{tabular}{l|c|c}
    Model & Factuality & Formality\\ 
    \hline
    Na\"{\i}ve Bayes & 0.50 & 0.78\\ 
    Logistic Regression & 0.51 & 0.78\\ 
    Support Vector Classification & 0.54 & 0.78\\
    \hline
    BERT (4 epochs, 2e-5 & 0.77 & 0.86\\
    learning rate, 32 batch size) & &\\
    BERT (2 epochs, 2e-5 & 0.79 & 0.86\\
    learning rate, 16 batch size) & &\\
    BERT (2 epochs, 2e-5 & 0.78 & 0.86\\
    learning rate, 32 batch size) & &\\
    BERT (2 epochs, 5e-5 & 0.79 & 0.86\\
    learning rate, 32 batch size) & &\\
\end{tabular}
\caption{Model Performance: The Macro Average $F1$ scores of different models on both tasks.}
\label{table2}
\end{table}
\begin{table}[t]
\centering
%\resizebox{.95\columnwidth}{!}{
\begin{tabular}{l|c|c|c|c}
    & Precision & Recall & $F1$ & $N$\\ 
    \hline
    Fact & 0.89 & 0.92 & 0.90 & 2,524\\ 
    Opinion & 0.76 & 0.70 & 0.73 & 856\\ 
    Neither & 0.78 & 0.72 & 0.75 & 282\\
    \hline
    Accuracy & & & 0.85 & 3,662\\
    Macro Avg. & 0.81 & 0.78 & 0.79 & 3,662\\
    Weighted Avg. & 0.85 & 0.85 & 0.85 & 3,662
\end{tabular}
\caption{Model Performance (2 epochs, 5e-5 learning rate, 32 batch size): Factual vs. Opinion-oriented vs. Neither.}
\label{table3}
\end{table}
\begin{table}[t]
\centering
%\resizebox{.95\columnwidth}{!}{
\begin{tabular}{l|c|c|c|c}
    & Precision & Recall & $F1$ & $N$\\ 
    \hline
    Informal & 0.87 & 0.79 & 0.83 & 1,393\\ 
    Formal & 0.87 & 0.92 & 0.89 & 2,085\\ 
    \hline 
    Accuracy & & & 0.87 & 3,478\\
    Macro Avg. & 0.87 & 0.85 & 0.86 & 3,478\\
    Weighted Avg. & 0.87 & 0.87 & 0.87 & 3,478
\end{tabular}
\caption{Model Performance (2 epochs, 2e-5 learning rate, 16 batch size): Formal vs. Informal.}
\label{table4}
\end{table}

When classifying sentences into ``factual'', ``opinion-oriented'', or ``neither of them'', the model (see Table \ref{table3}) reaches a Macro $F1$-score of 0.79 on the annotated test data and is best at predicting the ``factual'' cases, which could, to a great extent, be explained by the unbalanced amount of training data. This result is comparable to the English model obtaining a Macro $F1$-score of 0.76 in a previous study \cite{alhindi2020fact}. For the two-way classification (i.e., ``formal'' vs. ``informal''), the model (see Table \ref{table4}) achieves a Macro $F1$-score of 0.86 on the labeled test set and is relatively better at predicting the ``formal'' cases.

\subsection{Showcases}
\subsubsection{Showcase I.}

\begin{figure*}[t]
\centering
\includegraphics[width=1\textwidth]{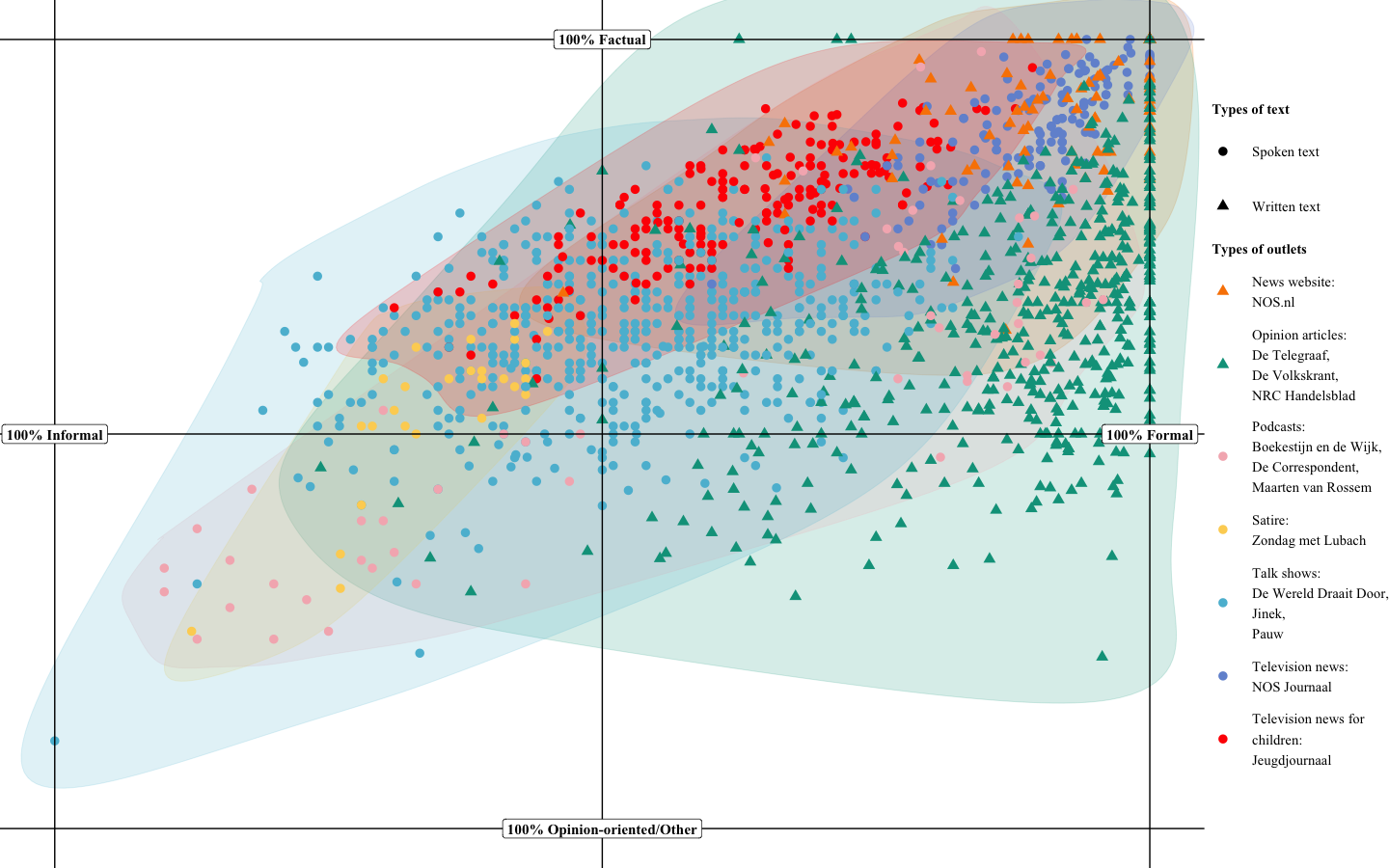} % Reduce the figure size so that it is slightly narrower than the column.
\caption{A two-dimensional grid with Dutch news items.}
\label{figure1}
\end{figure*}

\begin{figure}[t]
\centering
\includegraphics[width=0.46\textwidth]{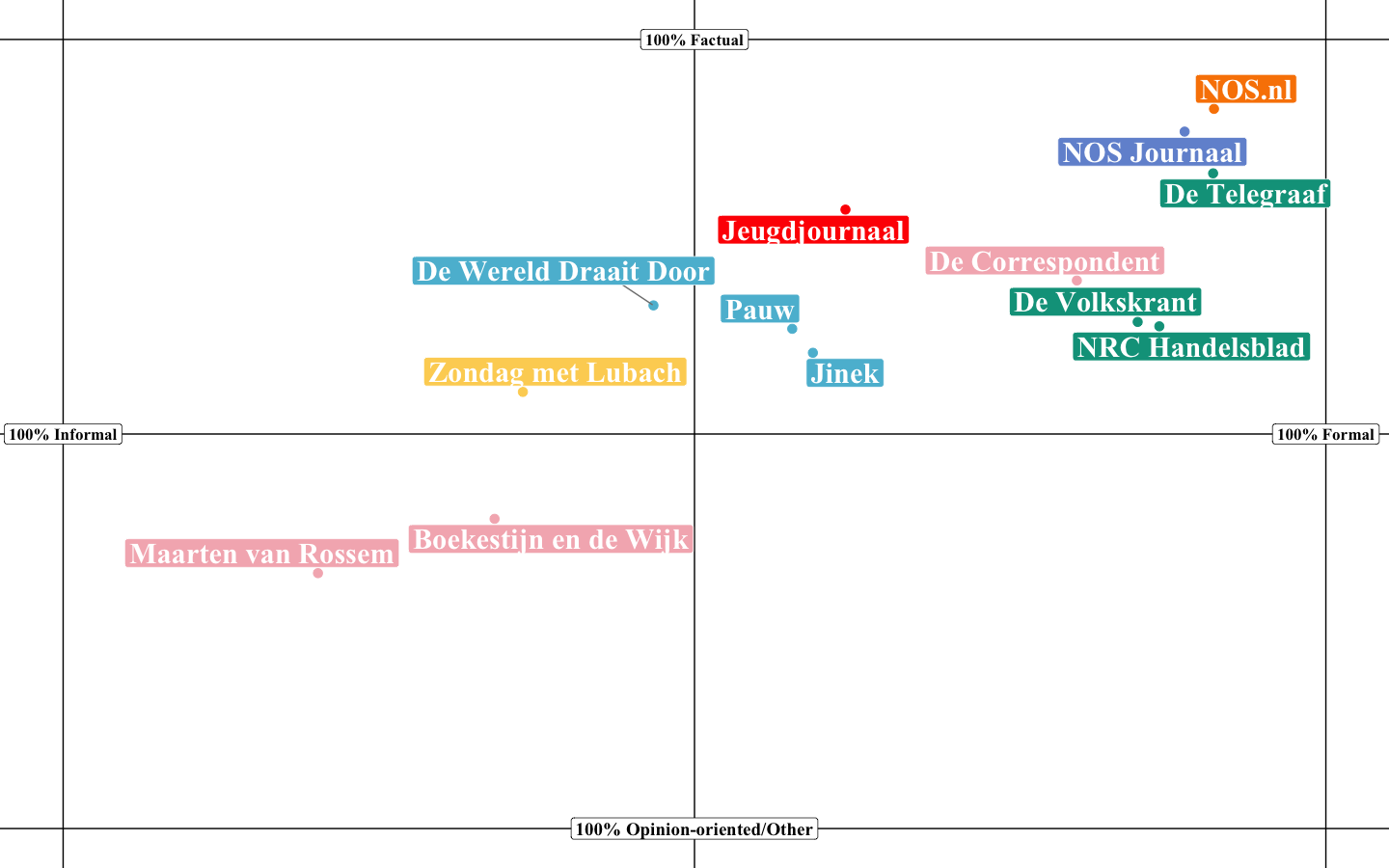} % Reduce the figure size so that it is slightly narrower than the column.
\caption{A two-dimensional grid with Dutch news items aggregated on the outlet level.}
\label{figure2}
\end{figure}

In this showcase, to capture how traditionally defined genres are positioned in our non-discrete framework, we collected 1,607 news items that cover both sources appearing in the training data and sources that have not been seen by the models (for source distribution see Table \ref{table5}). As the items are rather lengthy, we extracted the first 100 sentences for each item as representation, resulting in 113,427 sentences in total. This decision is also theoretically justified: we usually recognize the genre at the very beginning, not after finishing the whole item \cite[p. 6]{giltrow2009genres}. For illustration (see Figure \ref{figure1}), we first mapped the items on a two-dimensional grid where the position of each item indicates its percentages of different types of sentences on both scales. We colored the discrete genre groups and used different shapes to indicate the type of text for further comparisons.

\begin{table}[t]
\centering
%\resizebox{.95\columnwidth}{!}{
\begin{tabular}{l|l|c}
    & News Media Outlet & $N$ \\
    \hline
    Seen Sources & \textit{Boekestijn en De Wijk} & 12\\
    & \textit{De Correspondent} & 32\\
    & \textit{Jeugdjournaal} & 188\\
    & \textit{Maarten van Rossem} & 11\\
    & \textit{NOS Journaal} & 150\\
    & Opinion Articles & 439\\
    & \textit{Zondag met Lubach} & 26\\
    \hline
    Unseen Sources & \textit{DWDD} & 239\\
    & \textit{Jinek} & 151\\
    & \textit{NOS.nl} & 150\\
    & \textit{Pauw} & 155
\end{tabular}
\caption{Showcase I: An overview of the data (113,427 sentences from 1,607 items, 2008-2022).}
\label{table5}
\end{table}

Our grid (Figure~\ref{figure1}) is informative in different ways. First, it serves as a straightforward check for face validity: Items with spoken text (i.e., dots) are perceived as more formal by news consumers than the written-text items (i.e., triangles). More importantly, the colored shadows in the background indicate whether and how different discrete genres overlap with each other. The first impression is that the boundaries of traditionally defined genres are indeed permeable, justifying the need for moving beyond discrete genres. Looking closely, although the overlaps demonstrate that genres do not always limit to the exact same style, the stylistic patterns remain consistent within each genre. For instance, while the yellow shade of satire is included in the bigger blue shade of talk shows, signifying a similar presenting style, it is completely different from news websites and television news broadcasts.

To have a clearer comparison among different outlets, we aggregated the data points on an outlet level, resulting in Figure \ref{figure2}. Traditional news outlets such as \textit{NOS.nl}, \textit{NOS Journaal}, and newspapers (even though with opinion pieces) are positioned in the upper right corner, manifesting the typical definition of news --- being predominately factual and formal. The three blue talk shows occupying the central of the grid are slightly more factual and formal than the yellow satirical television programme. The red label, interestingly, shows that news for children is in general perceived as more informal and opinion-oriented than the regular news programmes. The pink podcast is the most peculiar genre, with the personal branding shows leaning towards the lower left quarter and the institutionalized news podcast lying on the upper right end of the spectrum.

\subsubsection{Showcase II.}
With this framework, we could also explore the differences within the same news outlet. The second showcase describes all news articles on \textit{NOS.nl} from 2017 to 2019 ($N = 64,612$), aggregated by the provided primary section/topic tags.

\begin{figure}[t]
\centering
\includegraphics[width=0.46\textwidth]{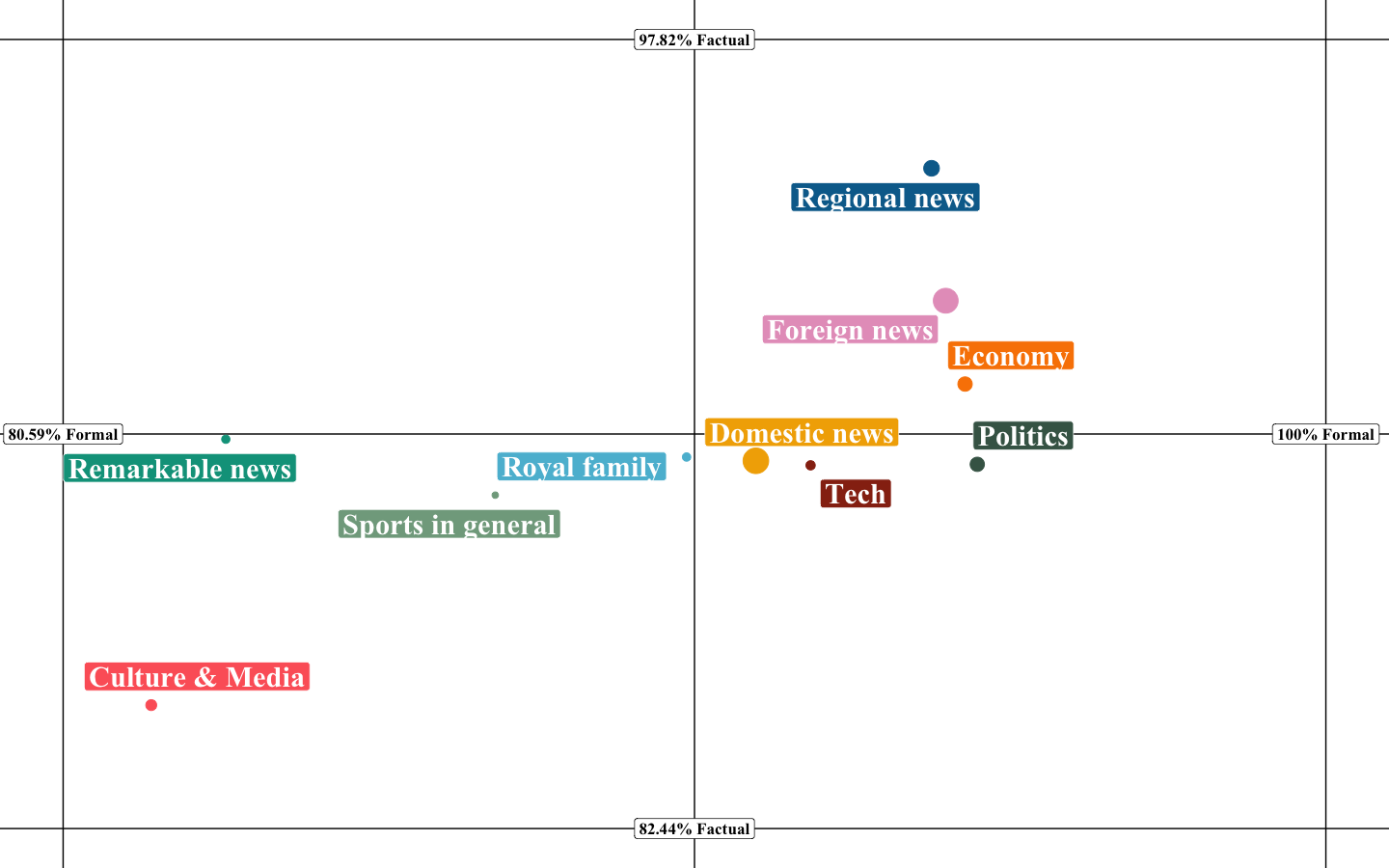} % Reduce the figure size so that it is slightly narrower than the column.
\caption{A two-dimensional grid with the \textit{NOS} data (zoomed-in grid: \begin{math}min./max.=mean\pm sd\pm5\end{math}). The size of the dots relatively represent the number of news items, with 235 sentences in the ``Sports in general'' section and 511,822 domestic news sentences. }
\label{figure3}
\end{figure}

Given that each item is quite short in length, all sentences were used for prediction, which are 1,328,977 sentences in total. With this considerable amount of data, the usual statistical tests would have lost their meaning, which is the reason that this study simply presents this descriptive grid with aggregated percentages to indicate the sectional diffidence and examine the fact validity. As Figure \ref{figure3} shows, the news articles on \textit{NOS.nl} are predominately factual and formal: Even the most informal and opinionated section (i.e., ``Culture and media'') has more than 80\% of the sentences being factual and formal. The ``Politics'' section, on the other hand, is even more formal and relatively more opinion-oriented compared to other sections. Other than that, the ``Sports in general'' section conveys more opinions while the ``Remarkable news'' section is more informal in style. 

\section{Discussion}

In this paper, to study news genres both in a continuum and from an audience's perspective, we theoretically developed a multidimensional framework for mapping news items in terms of genre cues. As a starting point, we operationalized it using two dimensions, namely ``factuality'' and ``formality''. For automatically obtaining the positions of either individual or aggregated news items within such a grid, we delivered a computational tool in the company of a large and various Dutch corpus with annotation. 

\subsection{Implications}

This multidimensional framework makes novel contribution from different perspectives, paving the way towards a suitable conceptualization in the contemporary media landscape. First, it is a framework of news genres. Acknowledging that the concept of news values has been overtly visible among the existing literature of communication science \cite{dent2008journalists}, the present study provides another angle for studying news by focusing on its forms and narrative styles \cite{broersma2007form}. Second, moving beyond discrete genre distinction, we describe news items in terms of the density of factual information and the formality level. It, therefore, avoids merely creating nominal categories, the typical pitfall of genre studies \cite{buozis2018reading}, and could accurately reflect the blurry and evolving nature of news genres \cite{steen2011feature}. Third, arguing that these two dimensions (i.e., ``factuality'' and ``formality'') could be perceived as indicative and picked up by news consumers during genre recognition, we refer to them as genre cues. This conceptualization enables us to add evidence estimating the audience's perception to the established newsroom-centered research. In addition, complementing the previous computational linguistic studies on news classification and formality detection \cite{alhindi2020fact, pavlick2016empirical}, this study sheds light on the ``demand-side'' with rather subjective annotations provided by actual news consumers \cite{costera2020understanding}.

Methodologically, this paper delivers a tool to operationalize this framework, where we automatically position a large amount of various news content, using the prediction from two BERT models as input for our visualizations. As presented by the above two showcases, the models’ prediction indeed corroborates the perception of general news consumers. In addition, our two-dimensional grid provides empirical evidence for our conceptualization that news genres are in flux but remain relatively stable. Although the non-random variations presented in our visualizations suggest that discrete genres are indeed reasonable, our approach allows more nuance comparisons on both individual and aggregated levels. This tool could be particularly helpful for communication scientists who intend to study news, especially a large amount of news, beyond the traditional definition and further draw inferences concerning how news genres have evolved.

Furthermore, this tool could be applied on different kinds of news items as long as the textual representation is available (e.g., articles in different forms, video subtitles, podcasts transcripts, posts on social media, etc.), regardless of spoken or written text. Although written text is naturally perceived as more formal than spoken text, showcase I indicates that they both widely spread across the grid, capturing a clear factual and formal distinction within both types. By adopting this tool, research questions of a various scopes could be informatively answered: For instance, cross-medium comparisons between newspapers and YouTube as we wondered at the very beginning of this paper, cross-platform observation concerning the same topic on Instagram and TikTok, or cross-outlet investigation into how news aggregators differ using search engine results. This tool is also helpful for detecting subtle patterns that are less intuitive to observe, such as how one single outlet varies in topics during the same period or changes in styles over time.

Practically, this study helps professionals better understand their audience, minimizes the perception gap between news production and news consumption \cite{costera2016revisiting}, and further improves journalistic decisions. Specially, we show the shifting and blending boundaries of news genres in a consumer's perception, enabling practitioners to have a sound grasp on how today's news genres have evolved and further adjust their mindset of defined genre separation \cite{buozis2018reading}.

\subsection{Ethics Statement}
For this study, we did not process any personal or otherwise sensitive data; we also did not in any way manipulate or mislead any human participant. Neither did we build a system that in any conceivable way leads to biases disadvantaging minority groups or similar. On the contrary, the release of our corpus could be beneficial to similar tasks in relevant disciplines, and our code could be easily applied by others. We expect no potential negative outcomes of these usages.

\subsection{Limitations and Future Research}
Conceptually, we chose two solid dimensions (i.e., ``factuality'' and ``formality'') as they are straightforwardly derived from the classic journalistic values. Yet, they are by no means exhaustive. We encourage future researchers to consider other potential dimensions as manifestations of genre cues as well. For instance, the density of humoristic elements could be indicative for consumers especially when recognizing satirical programmes \cite{otto2017softening}. Another potential dimension could be related to language complexity \cite{perez2017automatic}: As this concept is measured differently, one end of this dimension could be the level of readability, signaling news audiences the specific purposes for dissemination \cite{graefe2018readers}. Beyond linguistic features, framing is also relevant: Another non-discrete grid could be featuring the appearance of thematic versus episodic frames \cite{de2005news}, given that usually a news item does not exclusively fall into either one of these two categories \cite{iyengar1994anyone}. On the article level, the composition of an item is, too, insightful when it comes to genres \cite{keith2020evaluating,dai2021joint}, with institutionalized news pieces following the classic inverted pyramid structure. Moreover, although we currently focus on news consumption, the framework could be used for professionally defined concepts from the experts’ perspective (e.g., news editors, linguists, etc.).

On the other hand, albeit promising, our tool still has room for improvement. To begin with, though with a few YouTube videos included, the corpus overall does not contain a sufficient amount of social media content. Our candidate outlets are similar in a sense that each item tells at least one complete story in a flexible yet stable length consisting of at least several paragraphs, being greatly different from common social media posts. We argue that such a unique form of news deserves a model of its own. A more social media oriented model as the next step is therefore advised. Furthermore, aiming at training a comprehensive model, we constructed a general corpus with quite a mixture of news related data. As an interesting extension, future research could use the same approach but substitute our corpus with corpora in specific kinds. In this way, tailored models could better represent the layers within one kind. For example, a model specializing in fringe news outlets could more accurately capture the inherent variation within fringe news content than our general model \cite{bail2012fringe}. Building on that, it is also possible to draw comparisons between models separately trained with different types of news. For example, the same sentence might be labeled as informal by the mainstream model while formal by a fringe one.

Regarding the annotation, our corpus was not annotated on a crowd-sourcing platform but alternatively by a group of annotators, who are in general young and well-educated. A statement could be made that our models are better at predicting the perception of this specific demographic group of news consumers. It would again be interesting to compare models with different demographic input. For instance, whether the same sentence would be perceived as a fact or an opinion by models respectively trained with input from two politically polarized groups. As an improvement, future scholars could indeed consult a crowd-sourcing platform and obtain a more generalizable model on public perception \cite{benoit2016crowd}.

In a technical sense, considering the benefits of straightforward annotation tasks, our models were trained to perform classification on the sentence level. One could explore the otherwise regression approach to measuring the continuum by obtaining a continuous score for each sentence and later an average one for each item. Additionally, although the sentence-level decision allows us to flexibly apply the models to fragments with different lengths, which is not unusual considering the inclusive definition of news, an organic next step for future studies would be a document-level classifier with our models' output as the input. Similarly, the explained variance of these genre cues in predicting different item categories could provide solid justification for the dimensions chosen. Researchers are also recommended to train other kinds of models other than BERT using our released corpus, aiming for optimization.

In spite of the discussed limitation above, as the very first attempt to conceptualize and operationalize a non-discrete framework for mapping news items in terms of genre cues, this paper showcases promising output. In sum, our study provides scholars, practitioners, and news consumers a better understanding of the contemporary news ecosystem.

\section{Acknowledgments}
We thank our 17 annotators collectively for their help with the annotation.
\section{Funding}
This work is part of a project that has received funding from the European Research Council (ERC) under the European Union's Horizon 2020 research and innovation programme (Grant agreement No. 947695).
\section{Supplementary Materials}
To access our code and other supplementary materials, please visit: \url{https://github.com/zzzilinlin/beyond-discrete-genres}.

\bibliography{ref}
\end{document}